\documentclass[10pt,twocolumn,letterpaper]{article}
\usepackage[top=30mm,bottom=15mm,left=20mm,right=20mm,]{geometry}
\usepackage{times}
\usepackage{epsfig}
\usepackage{graphicx}
\usepackage{amsmath}
\usepackage{amssymb}
\usepackage{physics}
\usepackage{url}
\usepackage{amsthm}
\usepackage{algorithm}               
\usepackage{algorithmic}             
\usepackage{multirow}                
\usepackage{bm}
\usepackage{footmisc}

\def\1{\mathbf{1}}
\def\0{\mathbf{0}}

\newcommand\blfootnote[1]{%
  \begingroup
  \renewcommand\thefootnote{}\footnote{#1}%
  \addtocounter{footnote}{-1}%
  \endgroup
}
\newcommand{\rpm}{\raisebox{.2ex}{$\scriptstyle\pm$}}

\providecommand{\keywords}[1]{\textbf{\textit{Keywords:}} #1}

\begin{document}

\title{\LARGE \bf
MILDNet: A Lightweight Single Scaled Deep Ranking Architecture}

\author{Anirudha Vishvakarma$^{1}$\\
	$^{1}$Fynd (Shopsense Retail Technologies Pvt. Ltd.)\\ Mumbai, India.\\
	{\tt\small ~~anirudhav@fynd.com}\\  \vspace{-1mm}
	\blfootnote{The code is open-sourced at https://github.com/gofynd/mildnet}
}

\maketitle


\begin{abstract}\vspace{-3mm}
Multi-scale deep CNN architecture \cite{c1, c2, c3} successfully captures both fine and coarse level image descriptors for visual similarity task, but they come up with expensive memory overhead and latency. In this paper, we propose a competing novel CNN architecture, called MILDNet, which merits by being vastly compact (about 3 times). Inspired by the fact that successive CNN layers represent the image with increasing levels of abstraction, we compressed our deep ranking model to a single CNN by coupling activations from multiple intermediate layers along with the last layer. Trained on the famous Street2shop dataset \cite{c4}, we demonstrate that our approach performs as good as the current state-of-the-art models with only one third of the parameters, model size, training time and significant reduction in inference time. The significance of intermediate layers on image retrieval task has also been shown to be performing on popular datasets Holidays, Oxford, Paris \cite{c5}. So even though our experiments are done on ecommerce domain, it is applicable to other domains as well. We further did an ablation study to validate our hypothesis by checking the impact on adding each intermediate layer. With this we also present two more useful variants of MILDNet, a mobile model (12 times smaller) for on-edge devices and a compactly featured model (512-d feature embeddings) for systems with less RAMs and to reduce the ranking cost.
Further we present an intuitive way to automatically create a tailored in-house triplet training dataset, which is very hard to create manually. This solution too can also be deployed as an all-inclusive visual similarity solution. Finally, we present our entire production level architecture which currently powers visual similarity at Fynd.
\end{abstract}


\vspace{-4mm}
\section{Introduction}\vspace{-1mm}

\begin{figure*}[htp]
\centering
\includegraphics[width=\textwidth]{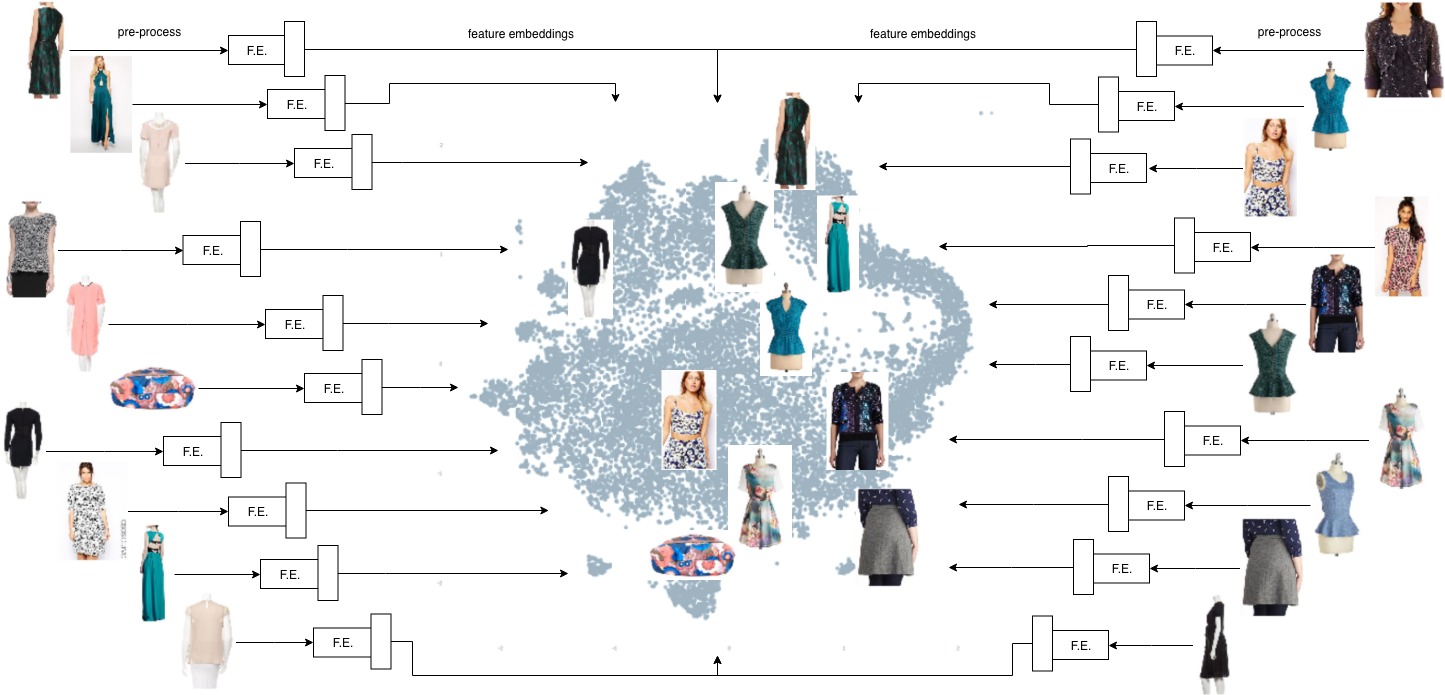}
\caption{Basic visual similarity pipeline where all the products in a database are mapped on an n-dimensional space using a feature extractor. Nearest neighbours are then the top most visual similar items.}
\label{fig:vs_basic_flow}
\end{figure*}

Instance-level-image retrieval (content based - CBIR) also known as visual similarity is a technique used for various tasks like visual recommendation, visual search, etc. and has proved to be a useful application and an active research topic for decades. The need of learning both fine level and coarse level abstractions complexly grained in the image pixel makes it a challenging task. 

Especially for an Ecommerce platform, where showcasing products in the right manner adds lot to the users' experience on product discovery. Hence, it is a critical feature to recommend users the products similar to what they are viewing as it readily captures their current intent (as opposed to similar users' taste using collaborative filtering or using past activities of user), leads to more engagement (CTR) and hence conversion (CR). For ecommerce, two basic use cases where CBIR is usually applied are:

1. \textbf{Visual Search}: Users upload a picture of someone who has worn the product (e.g. their friend or some celebrity) on the platform and a set of top visual similar items are presented from the entire dataset of the platform. Conversion rate could be very high if a good match is found this way.

2. \textbf{Visual Recommendation}: While user is browsing through any product on app, the top visual similar products to that product is readily displayed to user to capture his/her immediate interest. It can directly increase CTR.

Naively one can use text/meta details based search, but this data is not usually detailed and readily available. Also, the notion of similarity, especially for ecommerce domain, is not just a function of meta details of the product but is engraved in the visual appearance complexly present in the pixels of the product image. Capturing these minute details makes a big difference to the outcomes of such a feature. The task is not trivial as product comes in various variety even within the same class. Also, the notion of similarity is abstract, vague and debatable since it is usually relative. Let's take an example of a woman dress, the dress could be in variety of shapes (maxi, a-line, halter etc.), lengths (long, medium etc.), collar types (v neck, round neck, peter pan etc.), sleeve types (sleeveless, bell, puff, one-shoulder etc.), patterns (checked, striped, printed etc.), colors (red, maroon, pink etc.), etc. which the model should have an eye on. While the product image could be with or w/o a background and worn by a model of different complexion and hair colors in different poses, or a mannequin, or folded, or laid flat, are a few attributes our system should be ignoring. The solution hence is expected to understand fine-grained as well as coarse grained difference of details in two product images.

The visual similarity pipeline (see Figure ~\ref{fig:vs_basic_flow}) generally consists of a feature extractor which takes image pixels as input and produce a feature vector/embeddings representing the visual attributes needed for the task. The feature embeddings of all the images in the database when mapped to an n-dimension space places them in such a manner that visually similar products are always nearer. We can then use a technique like k Nearest Neighbour to quickly find k visual similar products to a query image.

\begin{figure}[htp]
\centering
\includegraphics[width=8cm]{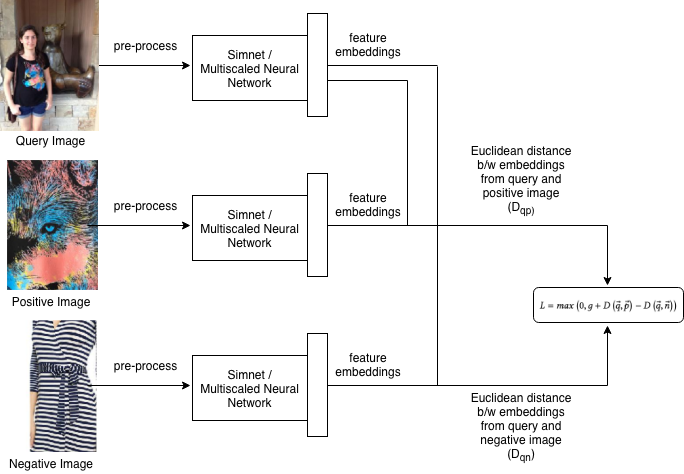}
\caption{Base Flow for triplets training. Each of the query, positive and negative images are passed through the CNN architecture. Embeddings are tuned to make sure that positive images are always closer to the query images than the negative in the latent space.}
\label{fig:triplet_training_flow}
\end{figure}

Handcrafted features (SIFT \cite{c12}, HOG \cite{c11}, etc.) have been tried as the feature extractor but failed to give high accuracy. A deep CNN is hence a suitable candidate as it can be expected to capture all these details while also being robust. Deep learning has achieved great feat in classification tasks, the features extracted from these trained networks are still found useful for other tasks including image retrieval. Using features from a single CNN trained on a classification task has been tried but fails to give high accuracy, due to the strong invariance encoded in its architecture. While training on classification task, a deep CNN generally encode strong invariance which grows higher towards the top layers, making it hard to learn the fine-grained image visual similarity. Current state of the arts uses an ensemble of 3 CNNs, also known as multi-scale architecture, first introduced by Jiang et. al \cite{c1}. Each CNN here is expected to capture different level of abstraction from the image. They are trained on image triplets containing a query image, a positive image (similar to query) and a negative image (relatively dissimilar to query image). Triplet training strategy is used where a triplet loss function is used to penalize the network whenever query image is closer to negative than positive images while training (see Figure ~\ref{fig:triplet_training_flow}). This enforces the network to learn relevant features to keep similar images together. The improvement in learning and results comes up with a trade-off of larger model size and latency on inference.

While multi-scale deep CNN architecture can learn complex relations required for visual similarity task, they also come up with high computation cost and latency. Since an image retrieval task might require generating thousands of image in real-time at production, it's great if the model can be light and fast as well. We counter the use of extra CNNs by arguing that a multi-layered CNN while convolving image or the feature maps of previous layers, already has been seen to automatically learn different level of detail (from edges, gradients, to eyes, nose, to faces) progressively at different layers \cite{c6}. The lower first few layers of a CNN captures more local patterns of objects like lines and corners, and as we go deeper, the later layers starts recognizing more complex features. Since multiple level of features are already present in multiple layers of a CNN, using only the feature from the last layer hasn't found the best utilization of its learning \cite{c7,c8,c9}.

We here present a novel deep CNN architecture, MILDNet (Multi-Intermediate Layers Descriptors Net), which uses only 1 CNN containing multiple skip connections to capture features from different layers of a single deep CNN model. Activations from a convolutional layers can be interpreted as local features describing particular image regions. We aggregated this local features inspired by \cite{c5,c7,c8,c9} to convert them into powerful global descriptors. Qili et. al \cite{c9} used a very similar approach to competitively perform in the Alibaba large-scale search challenge. We used global average pooling to aggregate features from each layer inspired by the work of Artem et. al \cite{c10}. Rigorous experiments and comparison with multi-scale network architectures mentioned earlier on popular Street2shop clothing similarity data \cite{c4} has shown that our model captures as good as features as current state-of-art, but with 3 times smaller network size (around 80 MBs) and inference latency. We used accuracy and recall as the metrics to compare. To validate the impact of each skip connections we progressively trained CNN with addition of each intermediate layer starting from no skip connection. We also have experimented multiple variants of MILDNet in which one variant reduces inference latency and network size further by roughly 3 times, while the other reduces the ranking cost by 2 times. Another variant is a mobile version is presented which trades-off 5.4\% top test accuracy to be around 11 times lesser in size than the current state-of-the-art models. This model can be deployed client side in an application to reduce the server load. Further we demonstrate a novel way to create in-house tailored catalog triplet data semi-automatically, which is hard to create manually. Since similarity is a factor of various attributes and relative similarity is rather subjective, using this approach can create a data directly from product catalog capturing one's notion of visual similarity. Coupling the architecture of MILDNet with the tailored training data from direct catalog can boost the overall performance for an ecommerce app. Finally, we explain our production deployment strategy and methods to optimize such a system. Hence, the contribution of this work is multi-fold and also comprehensive.

\vspace{-2mm}
\section{Related work}
\vspace{-1mm}
Since the task of image retrieval is not trivial, various studies are done in the past to solve it. The best results are obtained from Content-based image retrieval (CBIR) systems where distinctions are made based on the visual details of the image rather than the meta details. The basic approach is always getting embeddings from passing images pixels through a function and then searching for similar image in the embedding space. So, the main component most studied is hence the feature extractor.

Global features generalizes entire object (contour representations, shape descriptors, and texture features) are seen to be extracted using Histogram Oriented Gradients (HOG \cite{c11}). While local features describes the image patches (like the texture in image patch) has been extracted from SIFT \cite{c12}, OASIS \cite{c13} and local distance learning \cite{c14} learn fine-grained image similarity ranking models on top of the hand-crafted features. These hand-crafted features works fast but lacks expressive power.

Encoding hand-crafted features into bag-of-words (BoW) histograms had been a traditional approach \cite{c15}. A further compact representation is found to be achieved using Vector Locally Aggregated Descriptor (VLAD) \cite{c16} which achieves good results while requiring less storage. Other approaches like Fisher Vectors \cite{c18}, and, more recently, triangular embedding \cite{c19}, have also shown state-of-the-art for “hand-crafted” features like SIFT. 

Deep convolutional features have been used for image retrieval in various prior works. Razavian et al. \cite{c17} was among the first to investigate the use of CNN features for various computer vision tasks, including image retrieval. But the performance lagged behind that of simple SIFT-based methods which can be tackled by additionally incorporating spatial information. Qualitative examples of retrieval using deep features extracted from fully-connected layers have been studied \cite{c15, c20} which showed to outperform SIFT-like features. Gong et al. \cite{c21} improved it by introducing Multi-scale Orderless Pooling (MOP) where different fragments of image as passed through CNN and the activations from the fully-connected layer is aggregated by VLAD \cite{c16}. This introduced complexity of computing the full DCNN pipeline not only on the original image but also on a large number of multi-scale patches and further apply two levels of PCA dimensionality reduction. The works \cite{c22, c23} evaluated image retrieval descriptors obtained by the max pooling aggregation of the last convolutional layer. Artem et. al \cite{c10} showed that using sum pooling to aggregate features on the last convolutional layer leads to a much better performance. Hence, inspired by this work we chose global average pooling for aggregating the deep CNN features in our proposed solution.

While CNN features from last convolutional layer showed good results, but have less discriminability for instance-level-image retrieval since local characteristics of objects are not preserved. On the other hand features from intermediate layers have these local characteristics. Wan et. al \cite{c7} did a comprehensive study on applying CNN features to real-world image retrieval with model retraining and similarity learning. Encouraging experimental results show that CNN features are effective in bridging the semantic gap between low-level visual features and high-level concepts. Several cutting-edge studies \cite{c24, c25} suggested that mid-level features extracted from intermediate layers could obtain better performance than features extracted from the final layer. Joe et al. \cite{c8} experimented on all intermediate layers of GoogleNet and selected the best performant layer as the best representation of the image. However, this approach is inefficient to the large-scale image dataset because the ‘best’ layer of different objects may be various. Recently the work of Qili et. al \cite{c9} demonstrated great results using a novel representation
policy that encodes feature vectors extracted from different layers, called as ‘multi-level-image representation’. Our work is closely related to theirs but consists of experimentation on a different dataset, comparison with different architectures, loss function and aggregation step.

\begin{figure}[htp]
\centering
\includegraphics[width=8cm]{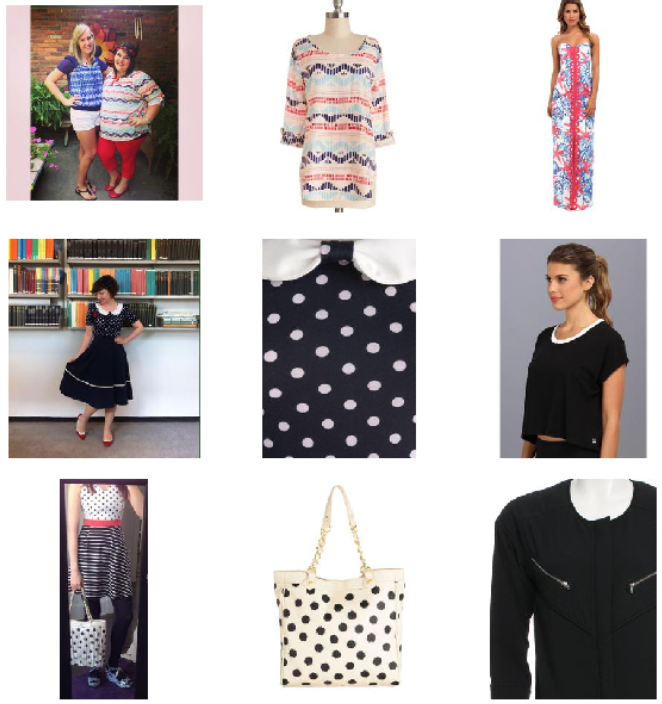}
\caption{Samples of triplets from Street2Shop data. Each row contains query, similar/positive and dissimilar/negative images. Query image is mostly a wild image, while positive and negative are catalog images. Last row is an example of out of class triplet pair.}
\label{fig:database_sample}
\end{figure}

Image similarity using Triplet Networks has been studied in \cite{c1, c26}. The work by Jiang et. al \cite{c1} introduced a novel multi-scale CNN architecture including Alexnet + 2 low resolution paths which demonstrated state-of-the art results for capturing both the global visual properties and the image semantics. A triplet-based hinge loss ranking function is used to characterize fine-grained image similarity relationships. Later Devashish et. al \cite{c2} improved the results further by using VGG16 instead of Alexnet. Recently, Rishabh et. al \cite{c3} showed state-of-the-art performance by using VGG19 as base convnet along with the use of a Siamese network with contrastive loss function \cite{c27}. In this paper, we will compare our model performance with these 3 works on the dataset ‘Street2Shop‘ which was made available by Kiapour et. al \cite{c4} (see Figure ~\ref{fig:database_sample}). This dataset contains curated, human labelled dataset containing wild/street image as query images and catalog/shop images as matching images. In most of the experiments we used triplet loss function which is a prediction error-based loss function, widely used for representation learning and metric learning in deep networks \cite{c1,c28,c29}. Our top performing model instead uses contrastive loss function \cite{c27} inspired by \cite{c3} which is rather a distance-based loss function.

\vspace{-2mm}
\section{Data Used}
\vspace{-1mm}

We have experimented our system on their ability of retrieving ecommerce product based on its visual appearance. For this we gathered triplet pairs consisting of a query image, a positive image (relatively similar to query image) and a negative image (relatively dissimilar to query image). The query image can either be
\vspace{-1mm}
\begin{itemize}
\item Wild Image: where people wearing the cloth in everyday uncontrolled settings.
\vspace{-2mm}
\item Catalog Image: model wearing cloth in controlled settings as shown in an ecommerce app.
\end{itemize}
While the positive and negative images can also be
\vspace{-1mm}
\begin{itemize}
\item In-class: same product category as query image
\vspace{-2mm}
\item Out-of-class: other product category than query image
\end{itemize}
Generally wild query images and both in-class and out of class positive and negative images are used for Visual Search feature where users can upload a picture of someone wearing a piece of cloth and find similar products from the entire catalog. While catalog query image with in-class positive and negatives are used for Visual Recommendation feature where on a product page a list of visually similar products are displayed to users in an ecommerce platform. So, for experimentation we used wild query image with catalog in-class and out-of-class positives and negatives. While for production deployment we used a mixture of both wild and catalog query image with majorly in-class positive and negative.
\subsection{Data for Experimentation}

We used Exact Street2Shop \cite{c4} dataset which is an extensive dataset which contains 20,000 wild images, 4,00,000 catalog images and exact street-to-shop pairs as meta-data (established manually). The retrieval sets is present for 11 fashion clothing categories, but only the category “tops” is rigorously experimented and presented here by us.  It also contains bounding boxes around the object of interest within the wild image. We sampled the query image, q of the triplet from the wild images (without cropping). Positive images, p of the triplet was always the ground truth match from catalog images. Negative images, n of the triplet were images of other catalog items. Finally, a training dataset of 70733 image and validation dataset of 25460 images is extracted and used.

\subsection{Data for Production Deployment}
\label{subsec:data_production}
For the production level model, the data should also consist catalog query images as that is what primarily will be queried in our app. Also, we wanted to make sure that the results should be tailored to our needs from visual recommendation system. So we decided to bootstrap our own data using an engineered way to tune the results (Section \ref{subsec:engineered_model}). Using results from this engineered model, we can now easily sample plenty of both in-class and out-of-class catalog triplet samples from our entire database. We used 30\% in-class to 70\% out-of-class triplets in case of catalog image triplets, while 30\% wild image triplets to 70\% catalog image triplets.

\vspace{-2mm}
\section{Approach}\vspace{-1mm}
\label{sec:approach}
We trained and evaluated our model as a triplet-based network architecture where 3 images (query q\textsubscript{i}, positive p\textsubscript{i} and negative n\textsubscript{i}) are passed independently into three identical deep ranking architectures with shared architecture and parameters. The triplets capture relative similarity notion of query image with positive image as compared to negative image. The system in this manner learns to create a feature embedding of dimension d, which is sufficient to capture both the fine-grained and coarse visual details of the image. During training the loss function makes sure that the resulting embeddings of query image is closer to positive images than the negative images. After getting these visual features, the problem now turn into a nearest neighbour search problem, where images with the closest embeddings to the query image gives visual similar items.
Hence the main components of this system is 1. Feature Extractor (CNN Architecture), 2. Loss Function (Hinge Loss), 3. Nearest Neighbour search (kNN).

\subsection{Architectures}
\subsubsection{Multi-scale deep ranking architecture}
Comparison of our proposed model MILDNet is done with 3 popular deep ranking architectures \cite{c1,c2,c3}. All the them are a type of a multi-scale neural network architecture first introduced by Jiang et. al \cite{c1}. The underlying architecture involves a combination of 3 separate CNN:
\vspace{-2mm}
\begin{itemize}
\item ConvNet which is usually Alexnet \cite{c15} and recently VGG16 \cite{c30}, VGG19 \cite{c30}. This CNN captures the image semantics and encodes strong invariance which can be harmful for fine-grained image similarity tasks. Also, the convnet is not trained from scratch and is pretrained on the popular ImageNet ILSVRC-2012 dataset \cite{c31}, which contains roughly 1000 images in each of 1000 categories.
\item Two shallower networks, receives down-sampled images to encode lesser invariance and capture the visual appearance.
\end{itemize}
The features from these three networks are concatenated and passed through a few fully connected layers to finally obtain a feature map which captures both the global visual properties and the image semantics.

\begin{figure}[htp]
\centering
\includegraphics[width=8cm]{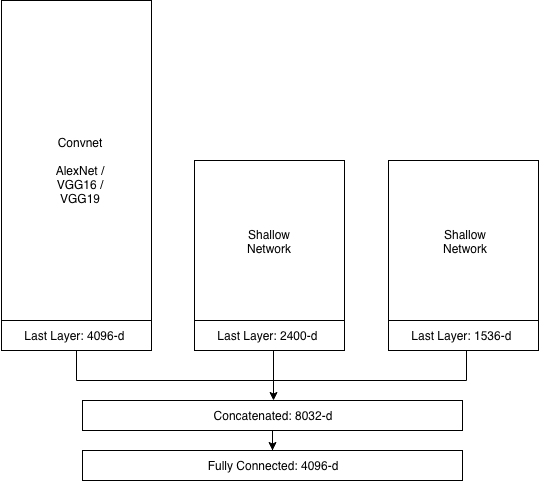}
\caption{Multi-scale network architecture consists of 3 independent CNNs: 1. Convnet(AlexNet/VGG16/VGG19) pretrained on ImageNet dataset 2. Shallow Network 3. Shallow network. The convnet captures the image semantics while shallow networks focus on the visual appearance. Embeddings from them are passed to a fully connected layer to get 4096-d embeddings.}
\label{fig:scaled_network}
\end{figure}

\subsubsection{MILDNet}
Rather than using complimentary CNNs to compensate the partial learning of the convnet in multi-scale architecture, we planned on using the features from few of the intermediate layers. As from many studies \cite{c6}, it is found that deep CNN capture different level of abstraction at different level of layers, this seemed intuitive. Our top performing variant of MILDNet uses VGG16 as the base Convnet, pretrained on ImageNet dataset. We extract features from 4 additional level than the last layer, just after the presence of max-pooling layers to limit the features extracted to most important ones. Artem et. al \cite{c10} have studied the performance of different aggregation methods for converting the CNN activations into global features. Inspired by their work, we used global average pooling to flatten the features and concatenate them to obtain a 1472-d feature vector. This is then passed through an FC - Dropout - FC layer to finally give the desired feature embedding of 2048 dimension. This way we significantly reduce the model size to the size of roughly a single VGG16 model. Also, while training we froze the first \textbf{10 layers} since the initial layers contains very local and generic features and need not be retrained.
\begin{figure}[hbt!]
\centering
\includegraphics[width=8cm]{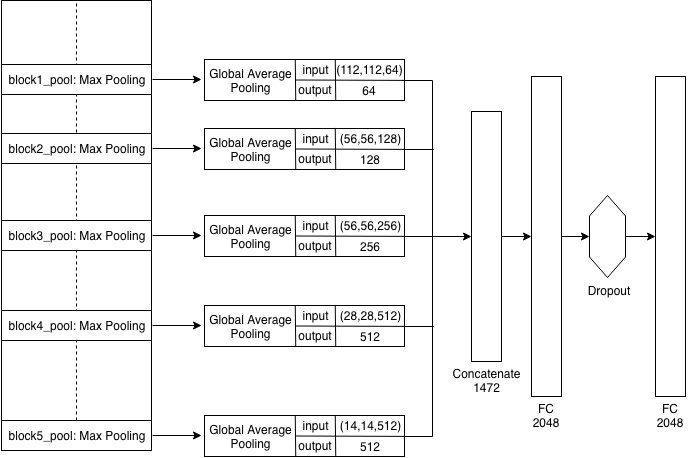}
\caption{MILDNet architecture only uses the Convnet(VGG16/MobileNet) pretrained on ImageNet dataset. Embeddings from multiple intermediate layers are aggregated on concatenate to get 1472 embeddings. This are passed through FC - dropout - FC layers to finally get 2048-d embeddings.}
\label{fig:MILDNet_arch}
\end{figure}

\subsubsection{Summary}
We tried out and compared five architectures in our experiments. Three of which are from recent research work on Multi-scale deep ranking architecture, and the rest 2 are variants of our proposed network, MILDNet. Table ~\ref{model_arch_comparison} shows the basic details of these five architectures. 
Finally, the feature vector can be summarized as a function f, which on passing input image I, outputs the image embedding $\Vec{x_i}$ in the embedding space.
\begin{equation}
\Vec{x_i} = f(I,W)
\end{equation}
where W contains all the weights and biases learned to optimize the mapping of images in embedding space.

\begin{table}[h]
\caption{Model Architecture Comparison}
\label{model_arch_comparison}
\begin{center}
\begin{tabular}{|c||c|c|}
\hline
Model & Size (MBs) & Total Params (M)\\
\hline
\hline
Ranknet\cite{c3} & 260 & 68.31\\
\hline
Visnet\cite{c2} & 253 & 66.54\\
\hline
Multi-scale-Alexnet\cite{c1} & 240 & 63.00\\
\hline
MILDNet(VGG16) & \textbf{83} & 21.93\\
\hline
MILDNet(MobileNet) & \textbf{20} & 5.33\\
\hline
\end{tabular}
\end{center}
\end{table}

\subsection{Loss}
Loss functions are the one which penalizes neural networks when they induce error while gradient descent tries to reduce the loss by adjusting the weights. In case of triplet image similarity where the base architecture have given 3 embedding feature vectors $\Vec{q}$, $\Vec{p}$ and $\Vec{n}$, loss functions make sure to keep the similar images together and dissimilar images apart in the embedding space. We tried two such loss functions: 
\subsubsection{Hinge Loss}
Hinge Loss function makes sure that in embedding space, vector $\Vec{q}$ is always relatively closer to $\Vec{p}$ than $\Vec{n}$ in terms of euclidean distance. Let's say that D($\Vec{x}$,$\Vec{y}$) denotes the euclidean distance of vector $\Vec{x}$ and $\Vec{y}$ in the embedding space, then the hinge loss function can be written as
\begin{equation}
 L = max(0, D(\Vec{q},\Vec{p})^2 - D(\Vec{q},\Vec{n})^2 + m)
\end{equation}
where m is the extra margin > 0, by which these vectors should be at least away and is decided empirically. We have used m = 1 in our experiments. 
As can be seen from the equation, when $\Vec{p}$ is farther to $\Vec{q}$ than $\Vec{n}$, the loss becomes positive by c plus the difference. This penalizes the model and weights are changed accordingly. Hence, hinge loss tries to keep the similar images closer relative to the dissimilar images.

\subsubsection{Contrastive loss function}
Contrastive loss function \cite{c23} is another distance-based loss function which also maps similar vectors to nearby points and dissimilar vectors to distant points. This loss function applies on a pair of samples rather like conventional learning systems which sums the loss over samples. The contrastive loss function can be written as
\begin{equation}
 L(\Vec{q},\Vec{p}) = 1/2*D(\Vec{q},\Vec{p})^2 
\end{equation}
\begin{equation}
 L(\Vec{q},\Vec{n}) = 1/2*(max(0, m - D(\Vec{q},\Vec{n})))^2
\end{equation}
where D is the euclidean distance between two vectors and m is the extra margin > 0 which is decided empirically. We have used m = 1 in our experiments.
When similar images are passed, the network is proportionally penalized by their distance. While for dissimilar images, the network is penalized only when distance is greater than the margin m. The paper also points that by simply minimizing the euclidean distance over the set of all similar pairs should lead to a collapsed solution. Contrastive Loss function have recently shown to have performed better than Hinge loss on the same task and dataset using multi-scale architecture \cite{c3}.

\subsection{Nearest Neighbour Search}
Finally, when we utilize the model to get a ranked list of visually similar images from the entire dataset, we need to search through the embedding space. We chose approximate nearest neighbour search for this task. The time complexity of exact Nearest Neighbour is O(n$^2$), which is reduced to O(n*k) by approximate nearest neighbours by allowing some errors. By definition, it works by finding a point p $\in$ P which is $\epsilon$-approximate nearest neighbor of the query q, that ∀p\textsuperscript{'} $\in$ P, d(p, q) ≤ (1 + $\epsilon$)d(p\textsuperscript{'},q). This definition says that instances which are only a factor of $\epsilon$ away from the real nearest neighbors can be considered as nearest neighbors.

We used Annoy (Approximate Nearest Neighbours Oh Yeah) \cite{c32,c33}, a library open-sourced by Spotify to aid us in this task. It is an algorithm based on random projections and trees. To compute the nearest neighbors it splits the set of points into half and does this recursively until each set is having k items where value of k is set empirically. To get better results, a priority queue sorted by the minimum margin for the path from the root node is used to search the tree. For every set multiple trees are build. If k items in the trees are found, duplicates are removed and for these instances the distances are computed on the original dataset.

\section{Experiment Details}
In all of our experiments, we used triplets extracted from the tops category of the famous Street2shop dataset, which finally gave us training dataset of 70733 image and validation dataset of 25460 images. We made sure that the triplets in validation dataset contains unique query images from the training dataset. These images are resized by their respective model architecture input size and rescaled by 1/255 to normalize. We chose Keras \cite{c34} with backend of Tensorflow \cite{c35} as the deep learning library. Training images are augmented using Keras's real-time augmentator, ImageDataGenerator class. We used following augmentations:
\vspace{-2mm}
\begin{itemize}
\itemsep-0.2em
\item horizontal\_flip: True, flips the image horizontally
\item vertical\_flip: True, flips the image vertically
\item zoom\_range: 0.2, zoom the image by scale of \rpm{0.2}
\item shear\_range: 0.2, shear the image by a factor of \rpm{0.2}
\item rotation\_range: 30, rotate the image by an angle of \rpm{30}$^{\circ}$
\item fill\_mode: nearest, fills extra pixels after the distortion on the basis of nearest points.
\end{itemize}
We kept a seed to make sure same augmentation applies in all experiments, so we can compare results and to make it reproducible.

To conceptualized our models on Google Colaboratory (also known as Colab), which is an open-sourced pre-configured ML playground. We ran all our experiments on Google Cloud ML Engine, which is a serive to train and deploy models at scale. Our experiments used one of 1. Single NVIDIA Tesla K80 GPU 2. Single NVIDIA Tesla V100 GPU 3. Cluster of four NVIDIA Tesla K80 GPUs 4.Single cloud TPU. We ran around a 100 training jobs, tried different architectures or combination of losses and other hyperparameters, out of which top 8 top performing results of a few architectures are presented here. For optimizer, we have tried both RMSProp algorithm and stochastic gradient descent (SGD) with a momentum. The best result we got is by using SGD with a momentum of 0.9 and learning rate of 0.001. We used a batch size of 96 in most of the experiments.

For monitoring, we used Google Tensorboard as well as a very handy log and metrics monitoring tool Hyperdash ~\cite{c37}. For visualization of dataset and results, as well as for reporting results and plotting graph we used Google Colaboratory notebooks.

\section{Results \& Conclusion}
We present here the results of 8 of our experiments, each of which varies on the architecture used or the loss function. For evaluation, the major metrics for us was triplet test data accuracy and average inference time. The details of the underlying architectures and loss functions are presented in Section ~\ref{sec:approach}. The Table ~\ref{model_arch_comparison} shows the basic details of the 8 experiments we carried. Details of these 8 experiments are below:
\vspace{-2mm}
\begin{itemize}
\itemsep-0.2em
\item \textit{Multiscale-Alexnet}: Multiscale model with base convnet as Alexnet and 2 shallow networks. We couldn't find a good implementation of Alexnet on Tensorflow, so we used Theano to train this network.
\item \textit{Visnet}: Visnet Multiscale model with base convnet as VGG16 and 2 shallow networks. Without LRN2D layer from Caffe.
\item \textit{Visnet-LRN2D}: Visnet Multiscale model with base convnet as VGG16 and 2 shallow networks. Contains LRN2D layer from Caffe.
\item \textit{Ranknet}: Multiscale model with base convnet as VGG19 and 2 shallow networks. Hinge Loss is used here.
\item \textit{MILDNet}: Single VGG16 architecture with 4 skip connections
\item \textit{MILDNet-Contrastive}: Single VGG16 architecture with 4 skip connections. Contrastive Loss is used here inspired by \cite{c3}.
\item \textit{MILDNet-512-No-Dropout}: MILDNet: Single VGG16 architecture with 4 skip connections. Dropouts are not used after feature concatenation.
\item \textit{MILDNet-MobileNet}: MILDNet: Single MobileNet architecture with 4 skip connections.
\end{itemize}

\begin{figure*}
\centering
\begin{minipage}{.5\textwidth}
  \centering
  \includegraphics[width=8cm]{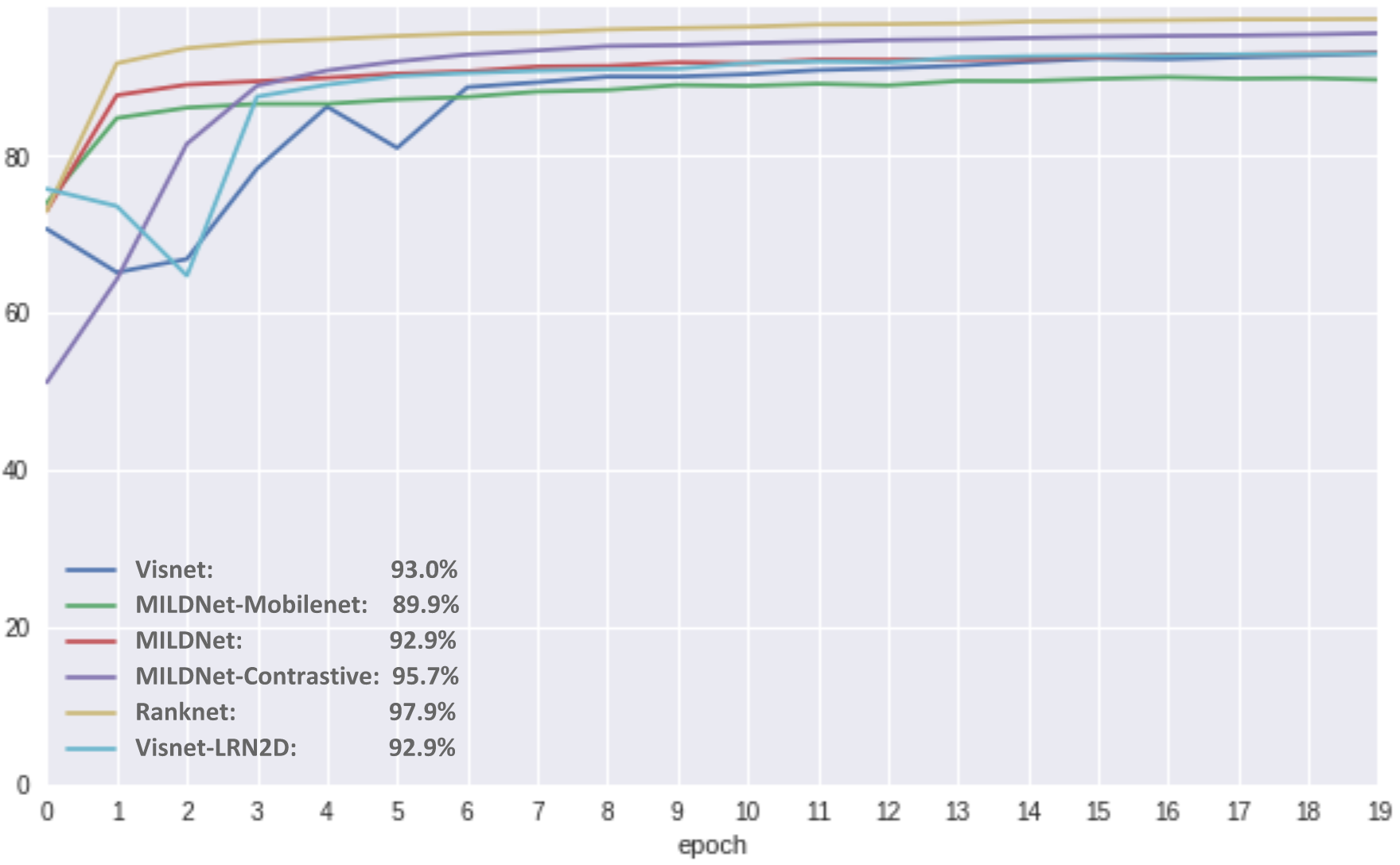}
  \caption{Trend of training triplet accuracy for different models.}
  \label{fig:training_accuracies}
\end{minipage}%
\begin{minipage}{.5\textwidth}
  \centering
  \includegraphics[width=8cm]{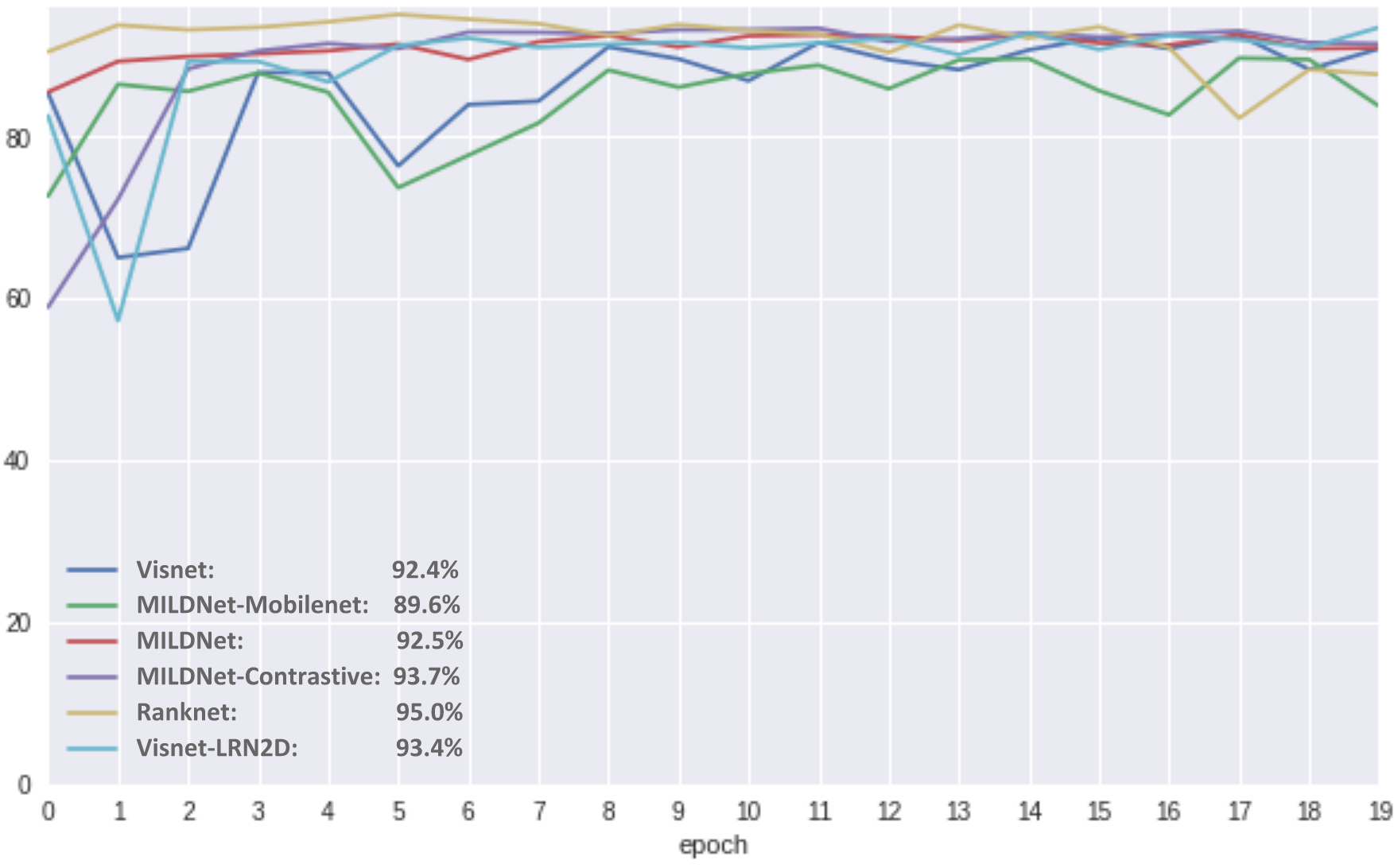}
  \caption{Trend of test triplet accuracy for different models.}
  \label{fig:test_accuracies}
\end{minipage}
\end{figure*}

Table ~\ref{mdoel_accuracy_results} shows training and validation triplet accuracy percentage of all the experiments, while the graphs ~\ref{fig:training_accuracies} and ~\ref{fig:test_accuracies} shows the trend of the training and validation triplet accuracy respectively. The test dataset consists of 25460 images consists of triplets from Street2shop dataset, having unique query images than the training data. Our model came second and only lags behind the top performing model (Ranknet) by only 1.29\% but with only one third of parameters and model size. This shows our model even being light has sufficient learning potential for this task. The model are further compared on term training and inference speed shown in Table ~\ref{model_performance_results}. A significant drop of training time and inference speed is shown by our proposed architecture. Further by using MobileNet architecture as base CNN instead of VGG16, we even brought down the inference speed to only 1ms by trading off the accuracy to 89.60\%. The training is done on a 24GB NVIDIA Tesla K80 GPU on Google Cloud ML Engine. The inference speed is tested on a 11GB GeForce GTX 1080 Ti GPU.

\begin{table*}[h]
\caption{Model Accuracy Results}
\label{mdoel_accuracy_results}
\begin{center}
\begin{tabular}{|c||c|c|c|}
\hline
Model & Loss Type & Max Training Accuracy (\%) & Max Test Accuracy (\%)\\
\hline
\hline
Ranknet & Contrastive & 97.89 & \textbf{94.98}\\
\hline
\textbf{MILDNet-Contrastive} & Contrastive & 95.69 & \textbf{93.69}\\
\hline
Visnet-LRN2D & Hinge & 92.87 & 93.39\\
\hline
MILDNet & Hinge & 92.94 & 92.50\\
\hline
Visnet & Hinge & 92.97 & 92.42\\
\hline
MILDNet-512-No-Dropout & Hinge & 92.8 & 91.15\\
\hline
Multiscale-Alexnet & Hinge & 87.72 & 90.80\\
\hline
MILDNet-MobileNet & Hinge & 89.90 & 89.60\\
\hline
\end{tabular}
\end{center}
\end{table*}

\begin{table*}[h]
\caption{Model Performance Results}
\label{model_performance_results}
\begin{center}
\begin{tabular}{|c||c|c|c|}
\hline
Model & Loss Type & \begin{tabular}{@{}c@{}}Avg. Training \\ Speed (mins/epoch)\end{tabular} & Avg. Inference Speed (ms)\\
\hline
\hline
Ranknet & Contrastive & 212 & 3.6\\
\hline
\textbf{MILDNet-Contrastive} & Contrastive & \textbf{85} & \textbf{2.6}\\
\hline
Visnet-LRN2D & Hinge & 197 & 3.2\\
\hline
MILDNet & Hinge & 88 & 2.6\\
\hline
Visnet & Hinge & 243 & 3.2\\
\hline
MILDNet-512-No-Dropout & Hinge & 91 & 2.6\\
\hline
Multiscale-Alexnet & Hinge & 165 & 3.0\\
\hline
\textbf{MILDNet-MobileNet} & Hinge & \textbf{70} & \textbf{1.0}\\
\hline
\end{tabular}
\end{center}
\end{table*}

We evaluated our models on test sets which were split apart from the complete dataset in the beginning. The test sets contain the same number of categories as that in the training set and also the class distribution was similar to that of the training data. The similar distribution also ensures that a generalized performance of the model is being measured.

\begin{figure}[htp]
\centering
\includegraphics[width=8cm]{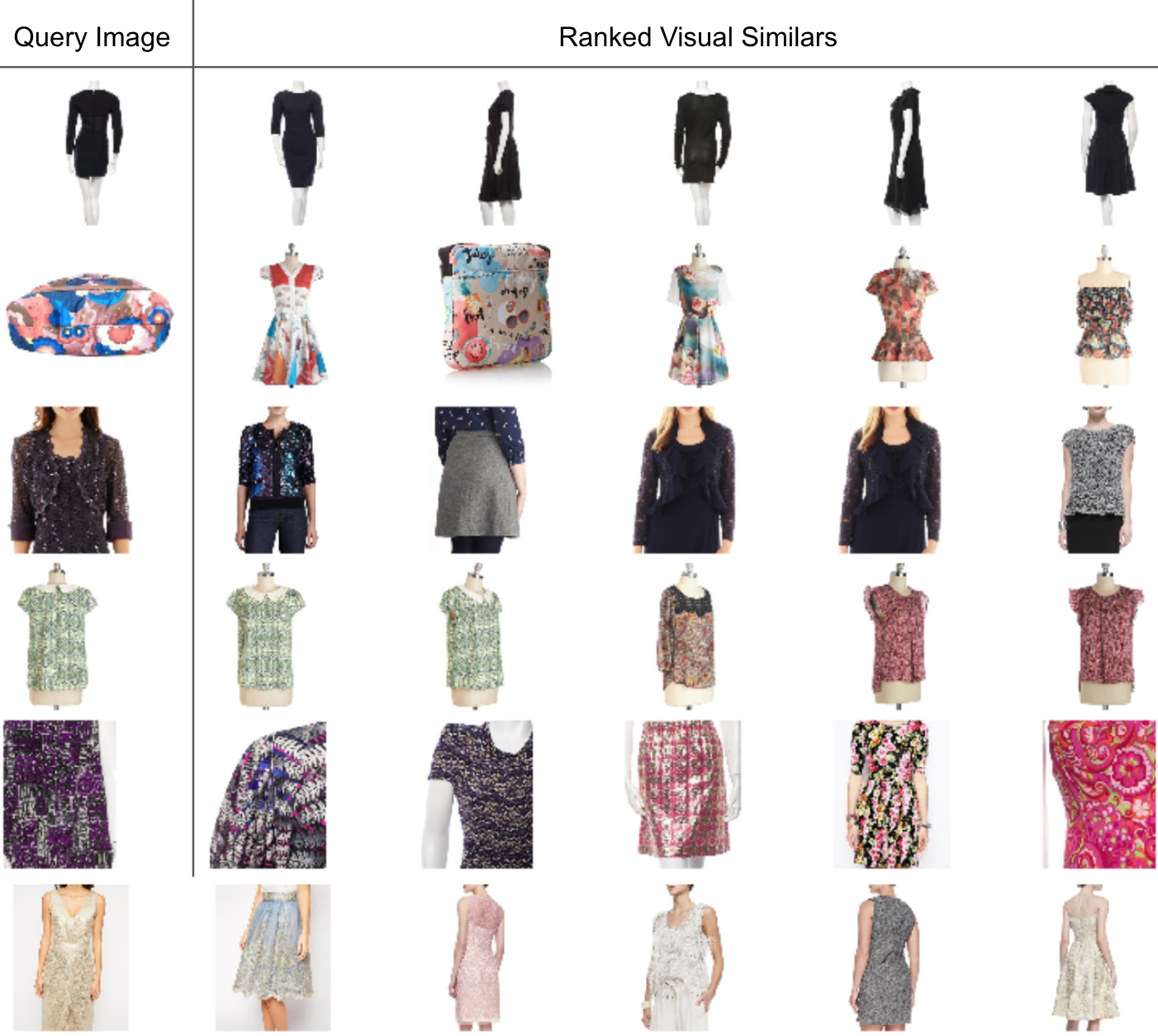}
\caption{Image Retrieval Samples of MILDNet. Note that since we only trained on tops category of Street2shop dataset, the second row doesn't contain only bag. But the model successfully shows understanding of both fine and coarse grained visual details.}
\label{fig:MILDNet_retrieval_samples}
\end{figure}

Finally, to ascertain the efficacy of skip connections, we did 4 more experiments where we started with no skip connection and gradually added aggregated embeddings from each intermediate layer as mentioned in the architecture of MILDNet (see Figure ~\ref{fig:MILDNet_arch}. Table ~\ref{MILDNet_ablation_study} shows the performance of these experiments. The results show a trend of increasing learning ability and inference accuracy with the contribution from each intermediate layer. For our training data we found that the second last layer "block4\_pool" contributed the most. But as been studied here \cite{c5}, the layer with the most contribution varies on the scale of input as well as the task. So to be comprehensive we considered all the 4 intermediate layers in our proposed solution.

\begin{table*}[h]
\caption{MILDNet Ablation Study}
\label{MILDNet_ablation_study}
\begin{center}
\begin{tabular}{|c||c|c|c|}
\hline
Skip Layers & \begin{tabular}{@{}c@{}}Total \\ Params(M)\end{tabular} & \begin{tabular}{@{}c@{}}Max Training \\ Accuracy(\%)\end{tabular} & \begin{tabular}{@{}c@{}}Max Test \\ Accuracy(\%)\end{tabular}\\
\hline
\hline
No skip & 19.96 & 70.35 & 71.15\\
\hline
block4\_pool & 21.01 & 92.45 & 91.05\\
\hline
block3\_pool, block4\_pool & 21.53 & 92.05 & 91.01\\
\hline
block2\_pool, block3\_pool, block4\_pool & 21.80 & 92.12 & 91.18\\
\hline
block2\_pool, block3\_pool, block4\_pool, block5\_pool & 21.93 & 92.94 & 92.50\\
\hline
\end{tabular}
\end{center}
\end{table*}

We also visualized the embeddings space obtained from our top performing model MILDNet with contrastive loss function. The 2048-d embeddings are projected to 2D using t-SNE \cite{c36}, which is a distributed stochastic neighbor embedding algorithm. The visualization showed that by training on Street2shop training data the model learned to recognize patterns and shapes present in the image. However, it failed to ignore backgrounds in the images which could be because query images were wild images while others were catalog images.

\begin{figure}[htp]
\centering
\includegraphics[width=8cm]{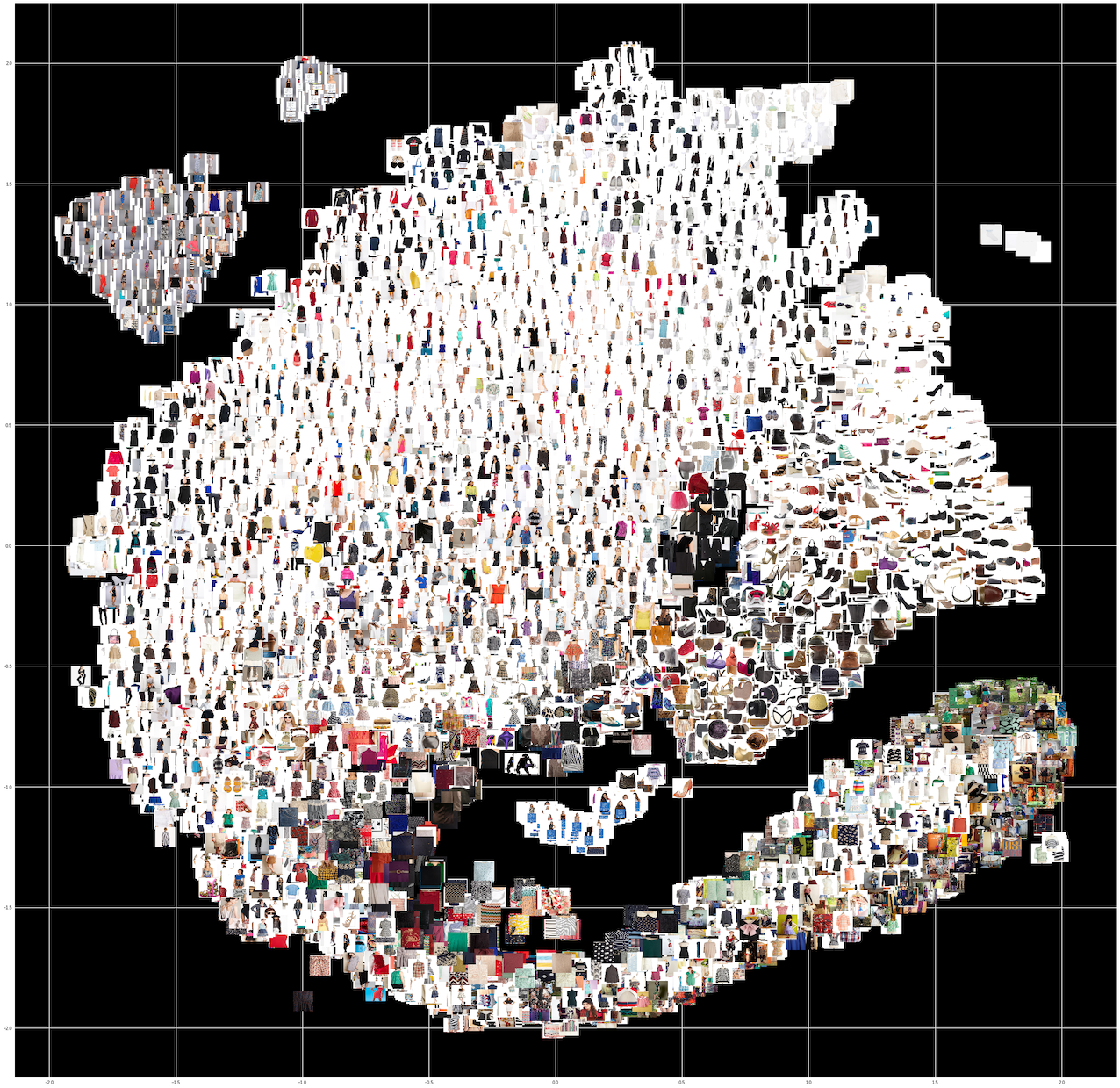}
\caption{t-SNE Visualisation of Street2Shop test data (only catalog images) made using features extracted from MILDNet.}
\label{fig:tsne_2}
\end{figure}

\vspace{-2mm}
\section{Production Details}\vspace{-1mm}
As discussed earlier in dataset creation for production (Section ~\ref{subsec:data_production}), we used both wild images triplets from street-to-shop dataset, and in-house automatically curated catalog image triplets from results of a unique and intuitive visual similarity model. Here we will first demonstrate our initial model which gave us enough catalog triplet data to train our final MILDNet model.
\vspace{-1mm}
\subsection{Configurable and Engineered Visual Recommendation Model}
\label{subsec:engineered_model}
\begin{figure}[htp]
\centering
\includegraphics[width=8cm]{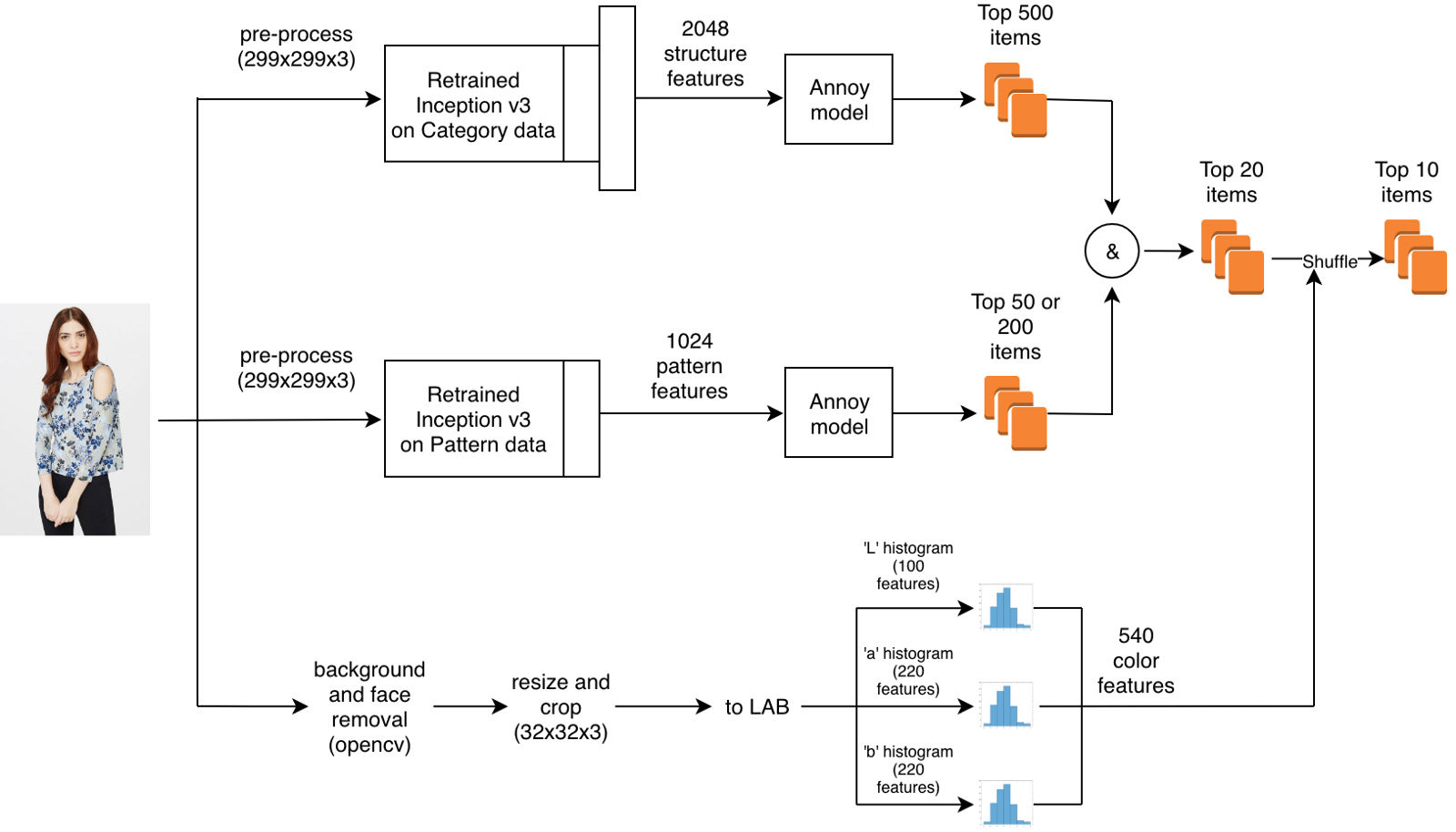}
\caption{Engineered Visual Recommendation Model. Features are extracted from pretrained 1. category classification model 2. pattern classification model 3. LAB colorspace histogram. Impact of these features on results are tuned to capture our notion of visual similarity.}
\label{fig:engineered_method}
\end{figure}

We identified three visual attributes which captures our notion of visual similarity:
\vspace{-2mm}
\begin{itemize}
\itemsep0em
\item Structure: the shape of the product
\item Pattern: the pattern/texture on the material of the product
\item Color: the primary color of the product
\end{itemize}
To capture these details from an image, we picked 3 deep classification models from our in-house models repository. Since, the second last layer of trained models contains automatically learned complex features necessary for the underlying classification task, we chose the following ways to extract hints of these features:
\vspace{-2mm}
\begin{itemize}
\item Structure: A product category classification model (42 categories like shirts, tshirts, sneakers, chinos etc.) retrained on InceptionV3 architecture. Feature vector from second last layer extracted containing 2048 features.
\item Pattern: A pattern classification model (7 pattern classes like solid, checked, melange, etc.) retrained on InceptionV3 architecture. Feature vector from the third last layer extracted containing 1024 features.
\item Color: LAB color space is used rather than RGB as they are closer resemblance to human perception. Histograms on LAB color space are stacked together to get color feature vector.
\end{itemize}
Using 2048 features representing structure, 1024 features representing pattern and 540 features from color histograms, we created kNN models using Annoy library. Finally, we created a pipeline (see Figure ~\ref{fig:engineered_method}) which we got after fine-tuning to get the results we desire.

This was our initial solution that we made live in a week span. The results from this solution is further used as training data after our experiments on MILDNet came positive. 
\vspace{-1mm}
\subsection{Infra Details}
\begin{figure}[htp]
\centering
\includegraphics[width=8cm]{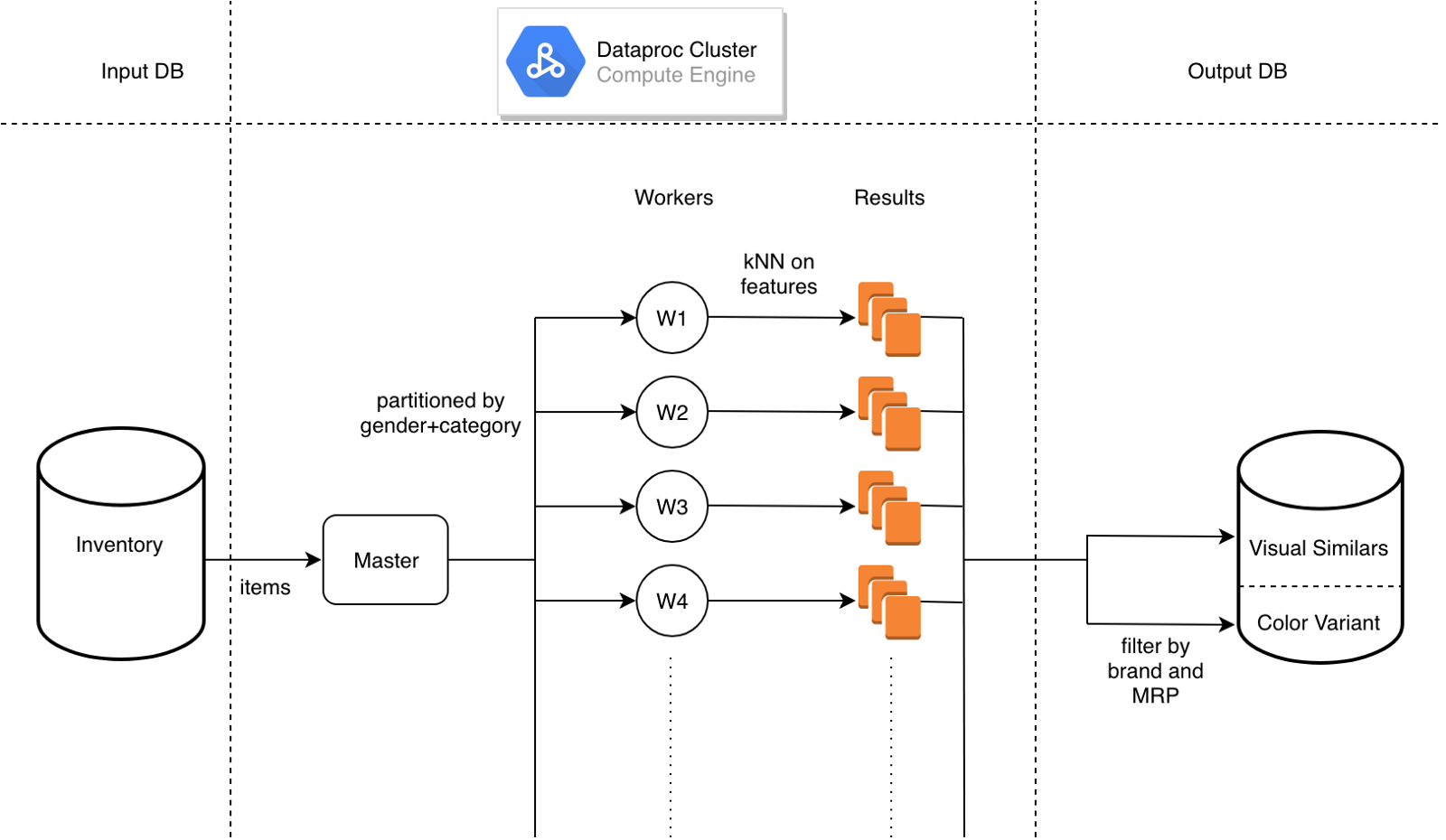}
\caption{The production flow of visual similarity pipeline at Fynd. Items from database are partitioned on gender+category keys and processed in batches on Spark cluster.}
\label{fig:production_flow}
\end{figure}

The Fynd catalog comprises thousands of products in around 500 categories, and each day around 10k insertions can happen. For this we decided to create a batch processing of data to find our visual similars of all products in the data once every day. Our system is roughly expected to first fetch each product, infer final feature vectors from either MILDNet or our initial model, then do a k nearest neighbour search to find top 10 visual similar products in the dataset. Our production setup (see Figure ~\ref{fig:production_flow}) takes roughly around 15 mins on let's say 6 worker clusters every day. It consists of following critical components:\vspace{-2mm}
\begin{itemize}
\item Databases: We use MongoDb database hosted on Amazon ec2 instances to store the product details (input db) and results (output db). 
\item Batching and Pruning search space: Since the database is huge and the results are only expected to be in-class, we decided to break it in batches. To cleverly create batches, we partitioned the entire dataset by two meta-data filed of the product, namely gender (men/women/girl/boy) and category key (last level of category among around 400 categories). This hugely eased the complexity and processing needed by the system by reducing the search space.
\item Cluster: To process the data in batches we used Google Cloud Dataproc clusters which works on Apache Spark framework in a distributed manner. The cluster consists of 1 high memory master, while number of workers are decided on run time based on the number of partitions to process.
\end{itemize}

\subsection{Optimizations}
Following are some of the optimization which are made possible by processing the data in partitions and also due to the high learning ability of MILDNet networks:
\vspace{-2mm}
\begin{itemize}
\itemsep0em 
\item Only those partitions are processed which has a new item added than the last run. 
\item Only those existing items are processed which are one among the nearest neighbour of new items.
\item Since the costliest process is getting features for new product images, the features once inferred are stored in a database for future requirements.
\item For mobile deployment, a variant of MILDNet where MobileNet with features from 5 intermediate layers is used instead of VGG16. We observed this to reduce the accuracy by around 5.4\%.
\item For real-time faster retrieval, we used a variant of MILDNet with only 512-d embedding instead of 1024-d embeddings by changing the number of features in last dense layer. We observed this to reduce the accuracy by only 2.1\%.
\end{itemize}

\section{Summary}
We have presented here a fresh take of using a single model with skip connections instead of using 3 CNN models to capture the notion of visual similarity. Experimenting on the famous Street-to-shop dataset we achieved equivalent accuracy and recall while reducing the model size and inference latency by 3 times. We also observed the effect of sequentially adding each skip connection. Different variants are also made with easy indexing using only 512-d final embedding and MobileNet variant with further 3 times lesser size (20 MBs). Since latency of such a system in live settings plays a key role, this could really boost the performance of such a system. Further we introduced a way to automatically create in-house tailored visual similarity catalog triplet data which is hard to create manually. Lastly, we demonstrated our entire production pipeline which caters batch processing entire ecommerce catalog dataset.

\addtolength{\textheight}{-10cm}

\end{document}